\documentclass[runningheads]{llncs}

\usepackage{graphicx}
\usepackage{amsmath,amssymb}
\usepackage{color}
\usepackage[width=122mm,left=12mm,paperwidth=146mm,height=193mm,top=12mm,paperheight=217mm]{geometry}

\usepackage{url}
\usepackage{xspace}
\usepackage{booktabs}
\usepackage{setspace}
\usepackage{tabularx}
\usepackage[noend]{algpseudocode}
\usepackage{algorithm,algorithmicx}

\usepackage[pagebackref=true,breaklinks=true,colorlinks=false,bookmarks=false]{hyperref}
\usepackage{cite}

\newcommand{\xbf}{\mathbf{x}}

\newcommand{\fzf}{\mathbf{z}}
\newcommand{\ybf}{\mathbf{y}}

\newcommand{\bPf}{\mathrm{P}}
\newcommand{\bEf}{\mathbb{E}}
\newcommand{\wif}{\mathbf{W}_I}
\newcommand{\wrf}{\mathbf{W}_R}
\newcommand{\wof}{\mathbf{W}_O}

\newcommand{\mcm}{\mathcal{M}}
\DeclareMathOperator*{\argmax}{arg\,max}

\makeatletter
\DeclareRobustCommand\onedot{\futurelet\@let@token\@onedot}
\def\@onedot{\ifx\@let@token.\else.\null\fi\xspace}

\def\eg{\emph{e.g}\onedot}

\def\etc{\emph{etc}\onedot} 
 
\def\etal{\emph{et al}\onedot}
\makeatother

\begin{document}
\pagestyle{headings}
\mainmatter
\def\ECCV16SubNumber{*}

\title{Learning Visual Storylines with Skipping Recurrent Neural Networks}

\titlerunning{Learning Visual Storylines with Skipping Recurrent Neural Networks}
\authorrunning{Gunnar A. Sigurdsson, Xinlei Chen, Abhinav Gupta}
\author{Gunnar A. Sigurdsson, Xinlei Chen, Abhinav Gupta}
\institute{Carnegie Mellon University\\\url{github.com/gsig/srnn} }


\maketitle

\begin{abstract}
What does a typical visit to Paris look like? Do people first take photos of the Louvre and then the Eiffel Tower? Can we visually model a temporal event like ``Paris Vacation'' using current frameworks? In this paper, we explore how we can automatically learn the temporal aspects, or storylines of visual concepts from web data. Previous attempts focus on consecutive image-to-image transitions and are unsuccessful at recovering the long-term underlying story. Our novel Skipping Recurrent Neural Network (S-RNN) model does not attempt to predict each and every data point in the sequence, like classic RNNs. Rather, S-RNN uses a framework that skips through the images in the photo stream to explore the space of all ordered subsets of the albums via an efficient sampling procedure. This approach reduces the negative impact of strong short-term correlations, and recovers the latent story more accurately. We show how our learned storylines can be used to analyze, predict, and summarize photo albums from Flickr. Our experimental results provide strong qualitative and quantitative evidence that S-RNN is significantly better than other candidate methods such as LSTMs on learning long-term correlations and recovering latent storylines. Moreover, we show how storylines can help machines better understand and summarize photo streams by inferring a brief personalized story of each individual album.
\end{abstract}

\section{Introduction}

In the past few years, there has been a remarkable success in learning visual concepts~\cite{chen2013neil,divvala2014learning} and relationships~\cite{chen2013neil,sadeghi2015viske} from images and text on the web. In theory, this allows the creation of systems that, given enough time and resources, can grow to know everything there is to learn. 
However, most of these approaches are still largely centered around single images and focus on learning static semantic relationships such as \textit{is-part-of}~\cite{chen2013neil}, \textit{is-eaten-by}~\cite{sadeghi2015viske} \etc. Moreover, many semantic concepts have not only a visual aspect but also a temporal aspect or even storylines associated with them. 
For example, a visual representation of \emph{Wedding} would involve guests entering the venue, followed by exchange of rings and finally celebrations in the wedding reception. How can we learn such visual storylines from the web as well?

\begin{figure}[t]
	\centering
	\includegraphics[width=0.8\linewidth]{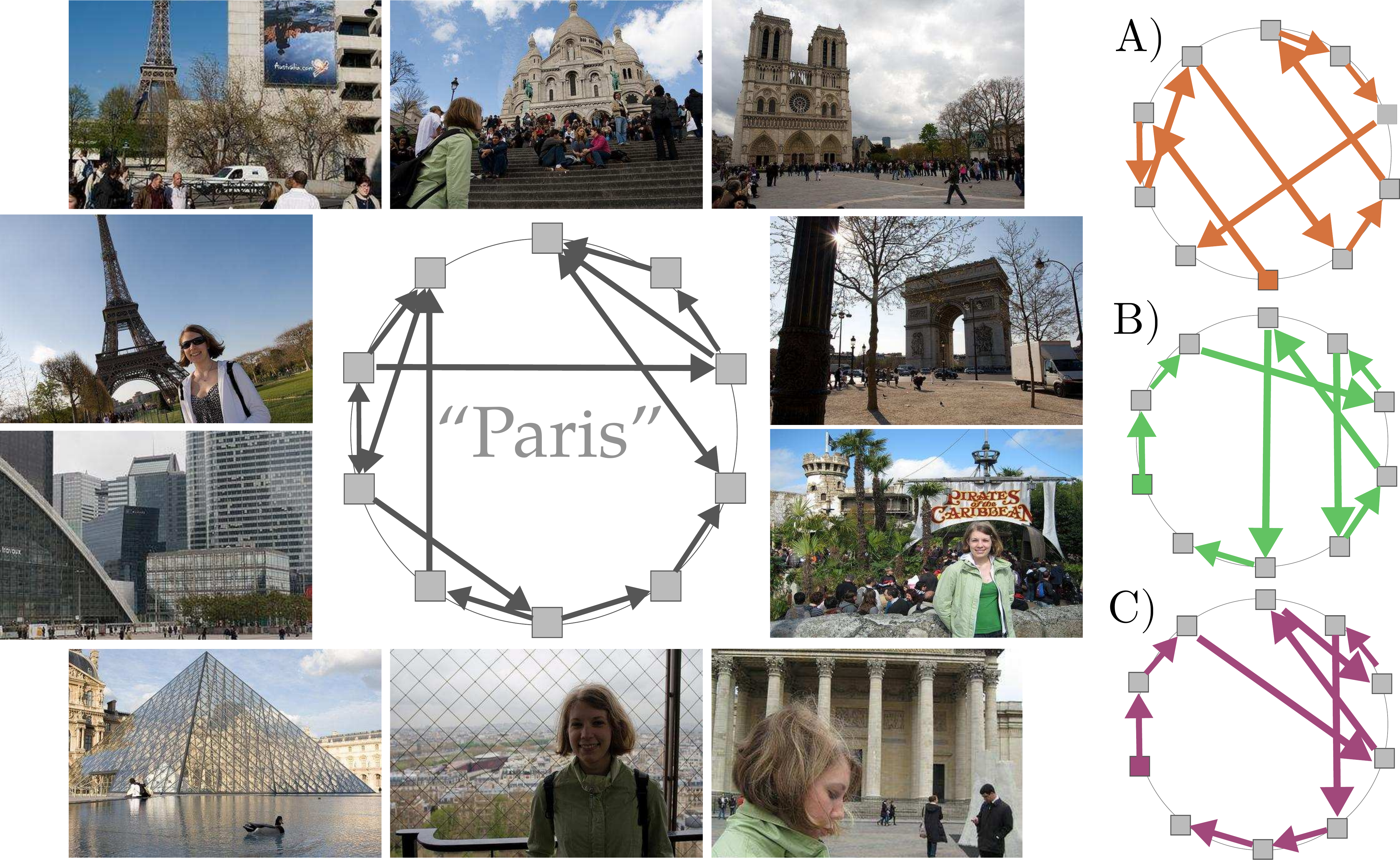}
	\caption{Given a concept, our algorithm can automatically learn both its visual and temporal aspect (storylines) from the web. To do this, we retrieve related albums from Flickr and apply our S-RNN model to automatically discover long-term temporal patterns. Here is a visualization of the storylines the model learned for the concept \emph{Paris}. For visualization, we distill the top images that a trained S-RNN model prefers by sampling storylines from a \emph{Paris} photo album.  Denoting the images as nodes in a graph, we visualize the %
	most common pairwise transitions using arrowed lines. On the right, we sample three probable storylines (A,B,C) that include these 10 images. We can see that the \textit{Eiffel Tower} is prominent early in the story followed by sightseeing of common landmarks (\textit{Arc de Triomphe} and others) and finally visiting the \textit{Lourve}. On a map of Paris, the \textit{Eiffel Tower} and the \textit{Arc de Triomphe} are indeed in close proximity  %
	\label{fig:teasernew}}
\end{figure}

\begin{figure}[t]
	\centering
	\includegraphics[width=1.0\linewidth]{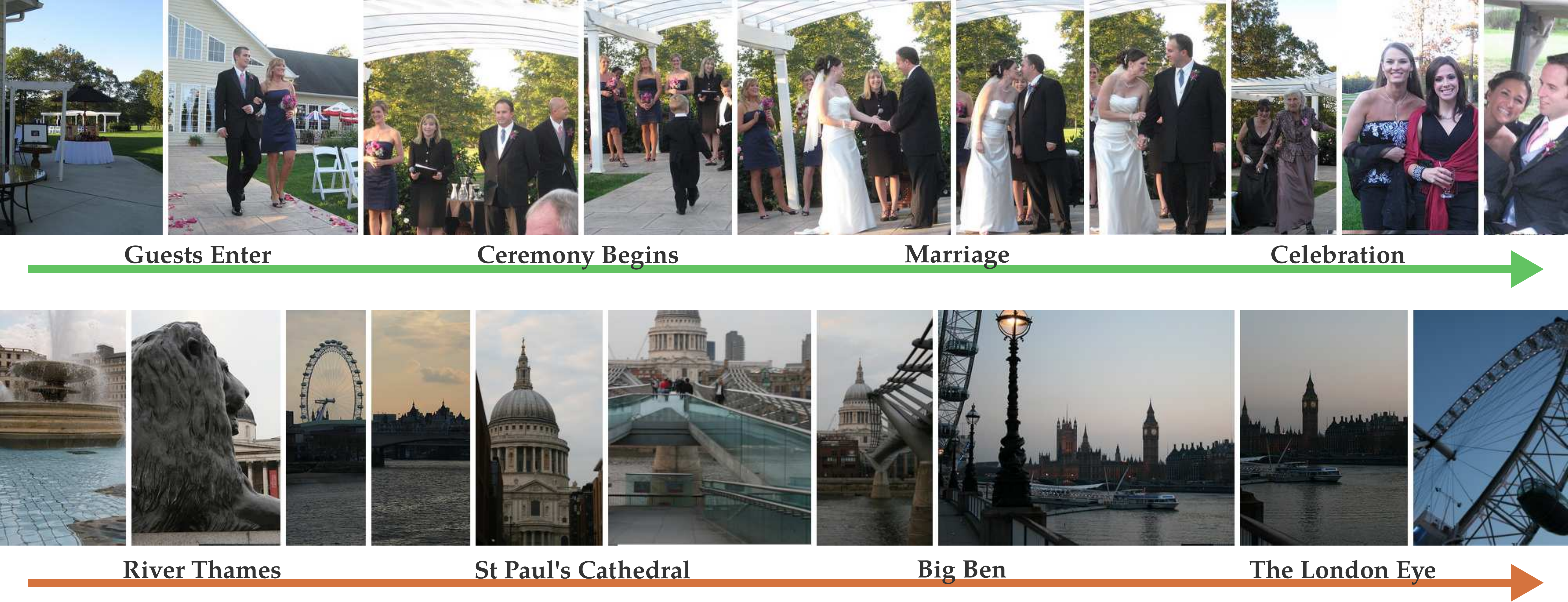}
	\caption{Given a concept, such as \emph{Wedding}, our algorithm can retrieve an ordered collection of images to describe that concept (Sec.~\ref{sec:imageretri}). In this figure we show the collections discovered by our model for two concepts. For example, for \emph{Wedding} (first row), it picks images that represent four steps: guests enter; ceremony begins, marriage and celebration. For travel-related concepts like \emph{London}, it prefers iconic landmarks for the story. The subtitles are manually provided for visualization. This is distilled from 1000 photo albums. More examples are provided in the appendix.\label{fig:teaser}}
\end{figure}

There are two aspects to these storylines: the visual aspect, often represented by modes in visual appearances, and the temporal aspect, which is the temporal order in which these modes appear. How do we capture both of these aspects from the web data? User photo albums in Flickr are a perfect example of web data that capture both aspects. First, most Flickr images are supplied with sufficiently informative tags, like \emph{Paris}~\cite{izadinia2014image}. Second, meta-information like time is usually available. In particular, the photos in each album are taken in ordered sequences, which hypothetically embed common storylines for concepts such as \emph{Paris}. Therefore, we propose to utilize Flickr photo streams across thousands of users and learn underlying visual storylines associated with these concepts.
What is the right representation for these storylines and how do we learn it?

Recently, there has been momentous success in using CNN~\cite{krizhevsky2012imagenet} features along with Recurrent Neural Networks~\cite{elman1990finding,malinowski2015ask,shih2015look,yang2015stacked,zhu2015visual7w,xiong2016dynamic} (RNNs) to represent those temporal dynamics in data~\cite{karpathy2014,donahue2014long,vinyals2014show,VenugopalanRDMD15,xu2015show,gregor2015draw,moviebook,chen2014learning}. 
We aim to extend that idea to modeling the dynamics in storylines. In theory, RNN can model any sequence, but has limited memory in practice, and can only learn short-term relationships due to vanishing gradients~\cite{bengio1994learning}.

Our Skipping Recurrent Neural Network (S-RNN) skips through the photo sequences to extract the common latent stories, instead of trying to predict each and every item in the sequence. This effectively alleviates the artifacts of short-term correlations\footnote{In our Flickr dataset, $71.1$\% of consecutive images are above average (cosine) similarity.} (\eg repetition) between consecutive photos in the stream, and focuses the learning effort towards the underlying story. This solution is complementary to, and different from, more complex RNN architectures such as LSTMs~\cite{hochreiter1997long} that still focus on learning transitions between consecutive images.
Similar to clustering, the S-RNN model can be efficiently trained in an \emph{unsupervised} manner to learn a global storyline and infer a private story for each album. Different from most clustering techniques, S-RNN inherits the power of RNNs that can capture the temporal dynamics in the data. 

We evaluate the effectiveness of our storyline model by comparing the storylines with baselines. In addition we evaluate the storyline model on two applications: a) image prediction~\cite{gunhee2014storygraph,kim2014joint}; and b) photo album summarization~\cite{dementhon1998video,ngo2005video,khosla2013large,martin20143d}.
Constructing a convincing storyline for a concept of interest requires both visual and temporal aspects. Therefore, algorithms need to retrieve a diverse collection of images, with the right ordering among them. For \emph{image prediction}, we show that our model is particularly suited for discovering the long-term correlations buried under the short-term repetitions in Flickr albums, while other approaches do not. Finally in the \emph{summarization} task, the goal is to take images in a single photo album and select a small summary of those. A typical example is a series of photos, taken by a family on their visit to \emph{Paris}, visiting all the iconic landmarks, such as the Eiffel Tower. Classically, summarization is approached by collecting a dataset of videos/albums and their associated summaries generated by people~\cite{khosla2013large,Sadeghi2015,xiong2015storyline,kim2015joint,chu2015video}, in order to learn how to make a summary in a supervised way. This process is, however, considerably laborious. In this work, we specifically experiment with the hypothesis that a quality summary of an album can be constructed by exploiting the similarities across thousands of similar albums (\eg \emph{Paris}). Then a summary of the album is inferred by telling a personalized version of the story.

\noindent \textbf{Contributions.} a) We present a new way of approaching sequence modeling with RNNs, by exploring all ordered subsets of the data to avoid short-term correlations between consecutive elements in the sequence. b) We present the novel {S-RNN} architecture that efficiently implements this idea on web-scale datasets. c) We demonstrate that this method can learn visual storylines for a concept (\eg \emph{Paris}) from the web, by showing state-of-the-art results on selecting representative images, long-term image prediction, and summarizing photo albums.

\begin{figure}[t]
	\centering
	\includegraphics[width=1.0\linewidth]{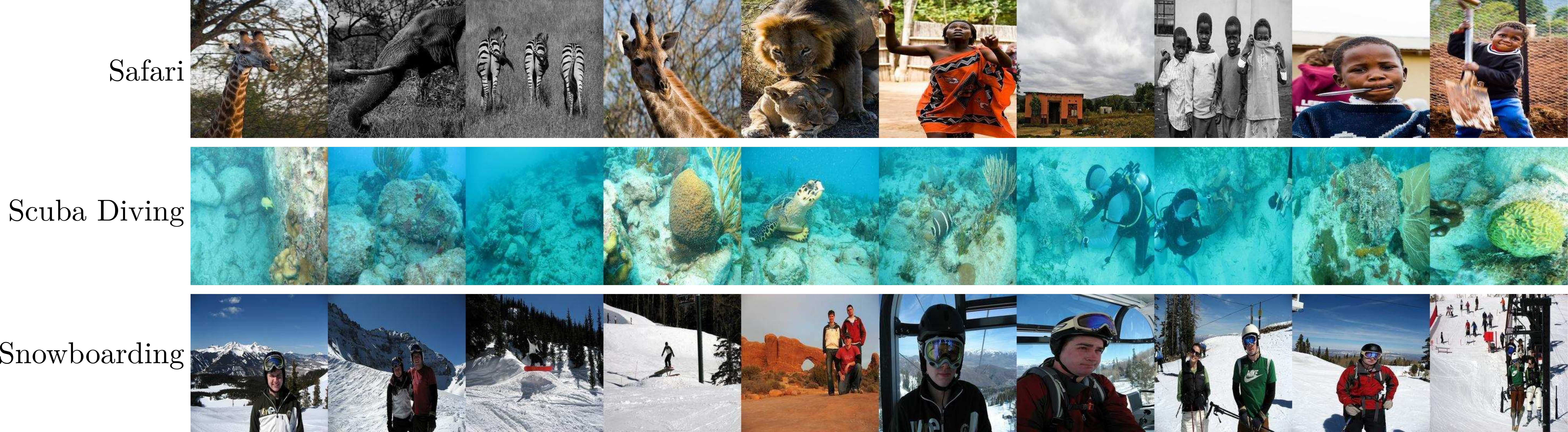}
	\caption{Given an individual photo album, our algorithm can summarize the photo album with a ordered collection of images that capture the album in terms of its underlying concept, by first learning about the concept from thousands of albums. (Sec.~\ref{sec:sum}). In this figure we show the summaries generated for three photo albums. One about a \emph{Safari}, the second about \emph{Scuba Diving}, and the third \emph{Snowboarding}. More examples are provided in the appendix.\label{fig:summaryteaser}}
\end{figure}

\section{Related Work}
\noindent \textbf{Learning storylines.} The earliest form of storyline can be traced back to the 1970-80s, where \textit{scripts}~\cite{shank1977scripts} (structured representations of events, causation relationships, participants, \etc) are used as knowledge backbones for tasks like text summarization and question answering. Unfortunately, these rich knowledge structures require hand construction by the experts, which fundamentally limits their usage in an open domain. This motivates the recent developments of \emph{unsupervised} approaches that can learn underlying storylines automatically~\cite{chambers2008unsupervised,mcintyre2009learning} from text. Inspired by this idea, our work aims to acquire the temporal aspect of a concept automatically from images. Similar work in vision is limited by either the scale of the data~\cite{wang2012generating,xiong2015storyline} or the domain to which the approach is applied~\cite{gupta2009understanding}. Perhaps the most similar work is~\cite{gunhee2014storygraph,kim2014joint}, where the storyline graphs are learned for Flickr albums. However, our work differs in several important aspects. First, while~\cite{gunhee2014storygraph} is an important step in learning storylines, it focuses its learning effort on each and every pairwise transition, but our method learns the long-term latent story. In fact, \cite{gunhee2014storygraph} could be extended using this framework, but here we extend a standard RNN model. Second, our method requires no a-priori clustering, feature independence, nor a Markov assumption, and does parameters sharing like RNNs.

\noindent \textbf{Temporal visual summarization.} Summarizing video clips is an active area of research~\cite{truong2007video}. Many approaches have been developed seeking cues ranging from low-level motion and appearances~\cite{dementhon1998video,ngo2005video,cernekova2006information} to high level concepts~\cite{lee2012discovering,khosla2013large} and attentions~\cite{ma2002user}. This line of research has been recently extended to photo albums, and more external factors are considered for summarization besides the narrative structure. For example, in~\cite{sinha2011summarization} the authors put forward three criteria: quality, diversity, and coverage. Later, in~\cite{obrador2010supporting} a system is proposed that considers the social context (\eg characters, aesthetics) into the summarization framework. Sadeghi \etal~\cite{Sadeghi2015} also consider if a photo is memorable or iconic. Moreover, most of these approaches are \emph{supervised}, namely the associated summaries for videos/albums are first collected by crowd-sourcing, then a model is learned to generate good summaries. While performance-wise it may seem best to leverage human supervision and external factors when available, practically it suffers serious issues like scalability and inconsistency in the ground-truth collection process, and generalizablility when applied to other domains. On the other hand, the task of summarization will be less ambiguous if the concept is given, which is exactly what we want to explore in this work. 

\noindent \textbf{Sequential learning with RNNs.} Recurrent neural networks~\cite{elman1990finding} are a subset of neural networks that can carry information across time steps. Compared to other models for sequential modeling (\eg hidden Markov models, linear dynamic systems), they are better at capturing the long-range and high-order time-dependencies, and have shown superior performance on tasks like language modeling~\cite{mikolov2010recurrent} and text generation~\cite{sutskever2011generating}. In this work we extend the network to model high dimensional trajectories in videos and user albums through the space of continuous visual features. Interestingly, since our network is trained to predict images several steps away, it can be viewed as a simple and effective way to learn long term memories~\cite{hochreiter1997long} and predict context~\cite{mikolov2013distributed} as well.
Fundamentally, LSTM still looks at only the next image and decides if it should be stored it in memory, but S-RNN reasons over all future images, and decides which it should store in memory (greedy vs.\ global). We outperform multiple LSTM baselines in our results. Furthermore, running LSTMs directly on high-dimensional continuous features is non-trivial, and we present a network that accomplishes that.

\section{Learning Visual Storylines}
Given hundreds of albums for a concept, our goal is to learn the underlying visual appearances and temporal dynamics simultaneously. Once we have learned this by building upon state-of-the-art tools, we can use it for multiple storyline tasks, and distill the explicit knowledge as needed, such as in Fig.~\ref{fig:teasernew}.
In this section, we explain our novel S-RNN architecture that is trained over all ordered subsets of the data, and show that this can be accomplished with update equations equally efficient to original RNN. The full derivation of these update equations by using the EM-method is presented in the appendix.
We formulate the storyline learning problem as learning an {S-RNN}. To understand S-RNN, we start by introducing the basic RNN model.

\subsection{Recurrent Neural Networks}
The basic form of RNN~\cite{elman1990finding} models a time sequence by decomposing the probability of a complete sequence into sequentially predicting the next item given the history (in our application, this sequence is images in a temporal order). 
Given a sequence of $T$ images $\xbf_{1:T}=\{x_1,\dots,x_T\}$,\footnote{For simplicity in notation, we assume a single training sequence, but in our experiments we use multiple albums for one concept to discover \emph{common} latent storylines.} the network is trained to maximize the log-likelihood:
\begin{align}
\mathcal{M}^* = \argmax_{\mathcal{M}} &\log \bPf(\xbf_{1:T};\mathcal{M}) - \lambda \mathcal{R}(\mathcal{M}) \nonumber \\
\mathrm{where}\ &\log \bPf(\xbf_{1:T};\mathcal{M}) = \sum_{t}{\log{\bPf\left(x_{t+1}|\xbf_{1:t};\mathcal{M}\right)}}.\label{eq:seq}
\end{align}
Here $\mathcal{M}$ is the set of all model parameters, and $\mathcal{R}(\cdot)$ is the regularizer (\eg $\ell_2$). 
The probability $\bPf(\cdot|\cdot,\cdot)$ is task dependent, \eg for language models it directly compares the soft-max output $y_t$ with the next word $x_{t+1}$~\cite{mikolov2010recurrent}. The standard optimization algorithm for RNNs is Back Propagation Through Time~\cite{williams1995gradient,werbos1988generalization} (BPTT), a variation of gradient ascent where the gradient is aggregated through time sequences.

The model consists of three layers: input, recurrent, and output. The input layer uses the input $x_t$ to update the hidden recurrent layer $h_t$ using weights $\wif$. The recurrent layer $h_t$ updates itself via $\wrf$ and predicts the output $\ybf_t$ via weights $\wof$. The update function at step $t$ writes as follows:
\begin{equation}
h_t=\sigma\left(\wif x_t + \wrf h_{t-1}\right);\ \ \ 
y_t=\zeta(\mathbf{W}_O h_t). \label{eq:y}
\end{equation}
Here $\sigma(\cdot)$ and $\zeta(\cdot)$ are non-linear activation functions, \eg sigmoid, soft-max, rectified linear units~\cite{krizhevsky2012imagenet}, \etc. All the history in RNN is stored in the memory $h_t$. This assumes conditional independence of $x_{t+1}$ and $\xbf_{1:t}$ given $h_t$.

In practice, the recurrent layer $h_t$ has limited capacity and the error cannot be back propagated effectively (due to vanishing gradients~\cite{bengio1994learning}). This can be a critical issue for modeling sequences like photo streams---due to the high correlation between consecutive images, where the dominant pattern in the short term is \emph{repetition}. For example, people can take multiple pictures of the same object (\eg the Eiffel Tower or family members), or the entire album is about things that are visually similar (\eg artwork in the Louvre or fireworks). This pattern is so salient that if an RNN is directly trained on these albums, the signals of underlying storylines are largely suppressed.
How to resolve this issue of learning long-term patterns? One way is to regularize RNN with a diversity term~\cite{sinha2011summarization}. However, note that if an album is indeed single-themed, we still want visually similar images in the storyline. Furthermore, Flickr tags are not perfect and noise in the album set can easily distract the model.

\subsection{Skipping Recurrent Neural Networks\label{sec:srnn}}
We now build upon the RNN framework to propose a skipping recurrent neural network model. Instead of learning each consecutive transition, S-RNN chooses to learn a ``higher-level'' version of the story, and focuses its learning effort accordingly. The key underlying idea is to select the storyline nodes by skipping a lot of images in the album and then modeling the transitions between the images selected as nodes. 

Formally, let us suppose $\xbf_{1:T}$ represents the $T$ images in the album, $\fzf_{1:N}$ is the set of indexes that represent the selected images for the storyline and the constant $N$ is the number of nodes in the storyline. Note that 
$N \ll T$, $z_n \in \{ 1,2,\dots,T \}$, and $z_n < z_{n+1}$ since $\fzf$ defines an ordered subset. Our goal is to learn the maximum likelihood model parameters ($\mathcal{M}$) by maximizing the marginal likelihood of the observed data. Therefore, our objective function is: 

\begin{align}
\label{eq:obj}
\mathcal{M}^* &=\argmax_{\mathcal{M}} \log \sum_{\fzf_{1:N}}{\bPf(\xbf_{1:T},\fzf_{1:N};\mathcal{M})} - \lambda \mathcal{R}(\mathcal{M}).
\end{align}

We can factorize ${\bPf(\xbf_{1:T},\fzf_{1:N};\mathcal{M})}$ as ${\bPf(\xbf_{1:T}|\fzf_{1:N};\mathcal{M})}{\bPf(\fzf_{1:N})}$ where ${\bPf(\fzf_{1:N})}$ is a prior on $\fzf$. As described above, we use a simple prior on $\fzf$ that it is an ordered subset. In this work, we make an assumption that the likelihood of a whole album is proportional to the likelihood of the selected sub-sequence of images $\xbf_\fzf$ (that is, we assume $\bPf(\xbf_{1:T} |\fzf; \mathcal{M}) \propto \bPf(\xbf_\fzf;\mathcal{M})$).
Factorizing, and inserting this assumption into Eq.~\ref{eq:obj} we have:
\begin{align}
\label{eq:obj2}
\mathcal{M}^* &=\argmax_{\mathcal{M}} \log \sum_{\fzf_{1:N}}\left(\prod_{n}{\bPf\left(\xbf_{z_{n+1}}|\xbf_{\fzf_{1:n}};\mathcal{M}\right)}\right){\bPf(\fzf_{1:N})} - \lambda \mathcal{R}(\mathcal{M}).
\end{align}
We observe that this equation is starting to look similar to standard RNN (Eq.~\ref{eq:seq}).

\begin{figure}[t]
	\centering
	\includegraphics[width=1\linewidth]{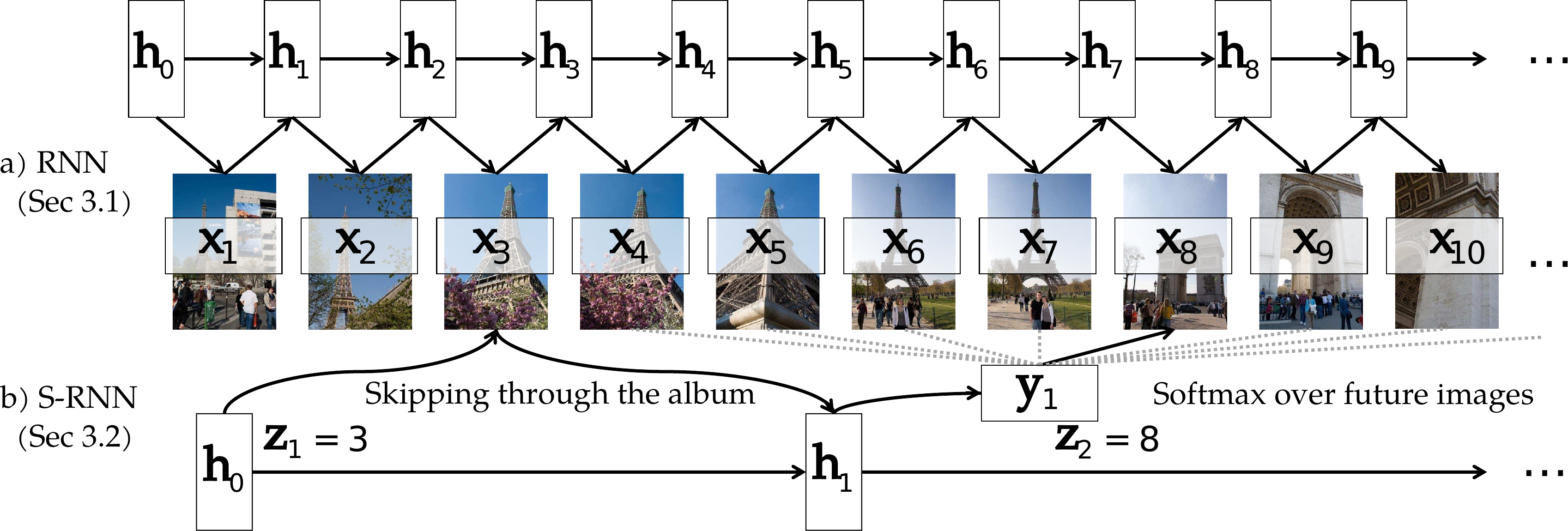}
	\caption{Our S-RNN model (unrolled in time). Instead of trying to predicting each and every photo in the sequence (as in the basic RNN model), latent variables $\fzf_{n}$ are introduced into our model to skip through the photo sequences, which is an effective strategy to address the local repetition issue (multiple pictures are taken for a single object like the Eiffel Tower) and can help extract common latent stories in the entire set of albums related to a concept (\eg \emph{Paris}). To overcome the high-dimensional regression problem, the loss is an softmax loss over future images \label{fig:model}}
\end{figure}

\subsubsection{Maximizing the S-RNN Objective}
Maximizing the marginal likelihood over all possible subsets of $\fzf$ is computationally intractable. Therefore, we make use of the Expectation Maximization (EM) algorithm, and then sequentially factor the update equations. More details of the EM derivation are given in the appendix. During the E-step, we sample $\fzf$ given the current model, and use that to train the model in the M-step, as we would an RNN. We initialize the EM-algorithm by setting $\fzf$ based on a randomly ordered subsets of images.

\subsubsection{S-RNN Implementation Details}
Now that we know how to optimize the objective, the only design choice left is the loss ${\bPf\left(\xbf_{z_{n+1}}|\xbf_{z_{1:n}};\mathcal{M}\right)}$ (the data likelihood in Eq.~\ref{eq:obj2}).
While Gaussian likelihood is often used for real-valued regression, we recognize that the space of allowed future images is not infinite, but simply images after $x_{z_{n}}$, defined as $\mathcal{X}_n$. Thus the likelihood is defined as a \emph{softmax likelihood over the future images}:
\begin{align}
    {\bPf\left(x_{z_{n+1}}|\xbf_{\fzf_{1:n}};\mathcal{M}\right)}
    = \dfrac{\exp (\ybf_n^T x_{z_{n+1}} ) }{
    \sum_{x \in \mathcal{X}_{n}} \exp (\ybf_n^T x )}
\end{align}
where $\ybf_n$ is the output of the network after step $n$. Effectively, this avoids modeling the negative world as ``everything except the ground truth'' and instead models the negative world as ``other possible choices''. This significantly helps with high-dimensional data ($fc7$ features), since the possible image choices in an album are usually only few hundred, but visual features few thousand.

In summary, during training and testing, $\fzf$ is sequentially sampled using the current model (which skips through the sequence), and during training those samples used to sequentially update the network with BPTT to maximize the objective. A visualization of the idea can be found in Fig.~\ref{fig:model}. A full implementation of is available.

\section{Experiments}
Since there has been so little done in the area of learning storyline models and their applications, there are no established datasets, evaluation methodologies, or even much in terms of relevant previous work to compare against. Therefore, we will present our evaluation in two parts: (a) first, in Section~\ref{sec:imageretri}, we directly evaluate how ``good'' our learned storyline model is. Specifically, we ask the Amazon Mechanical Turk (AMT) users how good our storyline model is compared to a baseline in terms of the representativeness and the diversity of image nodes in the storyline model; (b) next, in Section~\ref{sec:imagepred} and Section~\ref{sec:sum}, we evaluate our storyline model for two applications: long-term prediction and album summarization. For these tasks, we show qualitative, quantitative, and user studies to demonstrate the effectiveness of S-RNN based storyline model. We begin by describing our data collection process and the baselines.  

\subsection{Flickr Albums Dataset}
We gather collections of photo albums by querying Flickr through the YFCC100M dataset~\cite{thomee2015yfcc100m}, a recently released public subset of the Flickr corpus containing 99.3 million images with all the meta-information like tags and time stamps. This dataset is an unrefined subset of images on Flickr, making it a reproducible way of working with web data. The selection process gathers at most 1000 photo albums for a single concept (\eg \emph{Paris}), with an average size of 150 images. 
Each album is sorted based on a photo's date taken. We experimented with seven concepts: \emph{Christmas}, \emph{London}, \emph{Paris}, \emph{Wedding}, \emph{Safari}, \emph{Scuba-diving}, and \emph{Snowboarding} with a total number of 700k images. Examples from the dataset are provided in the appendix. This subset will be made available. 

\subsection{Implementation Details\label{sec:method}}
We compare our S-RNN model with several approaches to demonstrate its effectiveness in learning visual storylines. For fairness, all the methods used the same $\mathit{fc7}$ features from AlexNet~\cite{krizhevsky2012imagenet} pre-trained on ImageNet. 

For \textbf{S-RNN}, the $\mathit{fc7}$ features are directly fed into the model. The network is trained with BPTT, which unrolls the network, and uses gradient ascent with a momentum of 0.9. We set the starting learning rate as 0.05, and gradually reduce it when the likelihood on the validation set no longer increases. The input size of the layer is set to 4096 (size of $\mathit{fc7}$), and the hidden recurrent layer size 50. We keep $N=10$ for all the concepts as a good compromise between content and brevity (The appendix contains analysis of different sizes of N). We choose $\ell_2$ regularization and set weight decay $\lambda$ to be $10^{-7}$. Training takes approximately 2-3 hours on a single CPU. Each story was generated by sampling from the model 500 times, and picking the sampled sequence with the highest likelihood. The code is available at \url{github.com/gsig/srnn}.

Below we list the main baselines, and note that additional baselines will be added for individual experiments when necessary.

\noindent \textbf{Sample.} We uniformly sample from the data distribution.

\noindent \textbf{K-Means.} To take advantage of the global storylines shared in a concept, we apply K-Means to all the albums (similar to the first step of~\cite{gunhee2014storygraph} except with different features). 

\noindent \textbf{Graph.} We adapted the original code for~\cite{gunhee2014storygraph} to use $\mathit{fc7}$ features. 
Then a storyline is generated with the forward-backward algorithm as described in~\cite{gunhee2014storygraph}.

\noindent \textbf{RNN.} This architecture is similar to a language model~\cite{mikolov2010recurrent} except it predicts the cluster (as in \textit{K-Means}) of the next image. We sample without replacement to generate the story. This is a standard application of RNN to the problem.

\noindent \textbf{LSTM.} We train an LSTM network~\cite{KarpathyJL15}, similar to the RNN baseline.

\noindent \textbf{LSTMsub.} LSTM trained as before, but when generating the summary, we first generate a longer sequence ($N=100$) and then sub-sample that sequence to the desired summary length 10. Intuitively, if LSTM was indeed able to learn the long-term correlations regardless of the repetitions, this should perform well.

\noindent \textbf{S-RNN-.} For ablation analysis, we also provide a baseline where we use the network without skipping, but with the softmax loss over future images. All the hyper-parameters for training are kept identical to our model except the network predicts each and every item in the sequence. This is similar to RNN, but benefits from our improved loss.

\noindent \textbf{D-RNN} Similar to S-RNN-, except trained on a diverse subset of each album using the k-means++ algorithm~\cite{arthur2007k}. This was significantly better than other variants, including training on random subsets, fixed interval subsets, or a random diverse subset.

\subsection{Evaluating Storylines\label{sec:imageretri}}
\begin{figure}[t]
	\centering
	\includegraphics[width=1.0\linewidth]{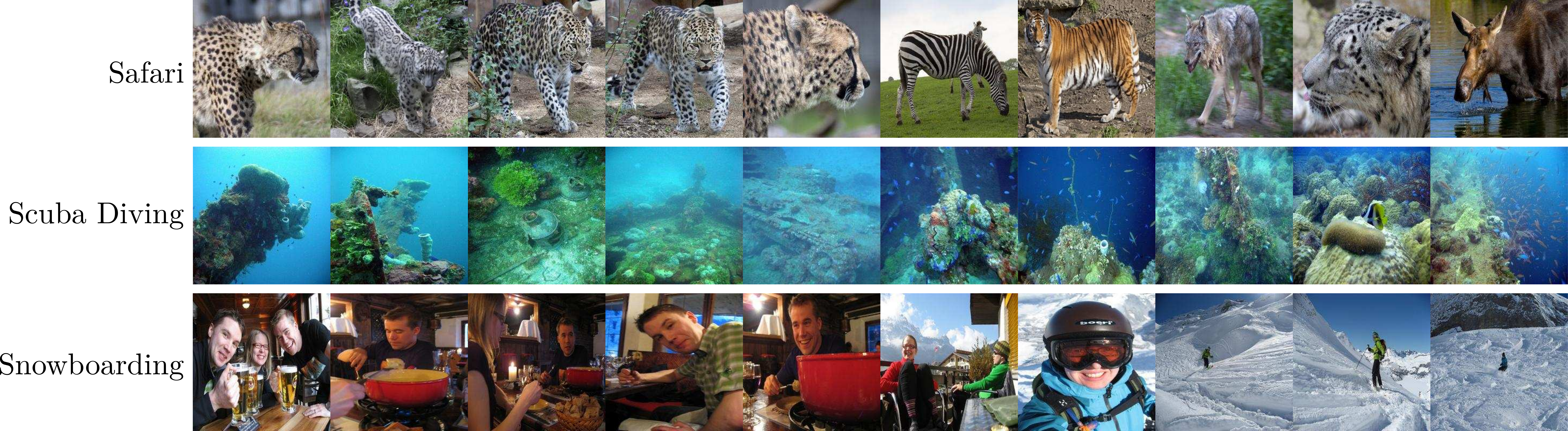}
	\caption{\emph{Evaluating Storylines}. Images selected by {S-RNN} for three storylines from thousands of images for the concepts \emph{Safari}, \emph{Scuba Diving}, and \emph{Snowboarding} \label{fig:storyexamples}}
\end{figure}

In the first experiment, we directly evaluate how ``good'' the learned storyline model is for a given concept.  We define the goodness of a storyline model in terms of how representative and diverse the selected images are for a given concept. Two qualitative examples for \emph{Wedding} and \emph{London} are shown in Fig.~\ref{fig:teaser}. Fig.~\ref{fig:storyexamples} shows more examples of learned storylines for different concepts. Our storyline model captures the essence of scuba-diving, snowboarding, etc., by capturing representative and diverse images (e.g., beer, fun and snowboarding during day).

\noindent \textbf{Setup.} For each concept, we have each method select only 10 images from 50 photo albums (thousands of photos) that best describe the concept, and AMT workers select which one they prefer. Each algorithm has access to the full training data to train the model. For Graph and RNN-based baselines, we sample multiple times from each album and use the highest ranked collection in terms of likelihood. Sample and K-Means are simply applied on all images, and in K-Means we assign the closest image to each cluster center. The appendix contains more qualitative examples.

\noindent \textbf{Results.} Table~\ref{tbl:retrieve} summarizes the results. Each comparison was given to 15 separate AMT workers. We can see that S-RNN is preferred \textbf{$\mathbf{60}$\% of the time against the strongest baseline} across all the concepts. Different baselines fail in different ways. For example, Sample and K-Means can capture a diverse set of images to represent the concept, but are prone to the inherent noise in the Flickr albums. On the other hand, Graph and LSTM overfit to the short-term correlations in the data and select repetitive images. Finally, S-RNN outperforms D-RNN since S-RNN is not restricted to a single specific diversity method as in D-RNN.

\begin{table}[t]
\centering
\caption{\emph{Evaluating Storylines}. Fraction of the time our S-RNN storylines are preferred against competing baselines. $50\%$ is equal preference. Our method significantly outperforms the baselines, being preferred \textbf{$60$\%} of the time against the strongest baseline. See Section~\ref{sec:imageretri} for details\label{tbl:retrieve}}
\setlength{\tabcolsep}{3pt}
\def\arraystretch{1.2}
\begin{center}
\begin{small}
\resizebox{\textwidth}{!}{%
\begin{tabular}{lcccccccc} \toprule
& {\bf K-Means} & {\bf Sample} & {\bf Graph} & {\bf LSTM} & {\bf LSTMsub} & {\bf D-RNN} & {\bf RNN} & \bf{S-RNN-}\\ 
\midrule
{\bf S-RNN} & 71.2\% & 68.3\% & 79.8\% & 84.3\% & 70.9\% & 60.0\% & 85.1\% & 75.5\% \\ \bottomrule
\end{tabular}
}
\end{small}
\end{center}

\end{table}%

\subsection{Task1: Prediction\label{sec:imagepred}}
Next, we evaluate our storyline models for two applications. The first application we consider is the {\em prediction} task. There are two possible prediction goals: short-term prediction and long-term prediction. Short-term prediction can be considered as prediction of the next image in the album. This was the task used in~\cite{gunhee2014storygraph,kim2014joint}. In the case of long-term prediction, we predict the next representative event. In the case of {\em Paris} vacation, if the current event is Eiffel tower, the next likely event would be visiting the Trocadero. In the case of {\em Wedding}, if the current event is the ring ceremony, then the next representative event is the kiss of the newlyweds.

\noindent \textbf{Setup.} For the short-term prediction, the ground-truth is the next image in the album. But how do we collect ground-truth for long-term prediction? We ask experts to summarize the albums (hoping that album summaries will suppress short-term correlations and capture only representative events). Now we can reformulate long-term prediction as predicting the next image in the human-generated \textit{summary} of the album. 
We collected 10 ground truth summaries on average for each concept from volunteers familiar with the concepts (such as \emph{Paris}, and \emph{London}). Each summary consists of 10 images from a photo album that capture what the album was about. This was used as ground truth only for evaluation.
Two settings are compared, the first one (labeled ``\emph{long-term}'') predicts the next image in a \emph{summary} ($N{=}198$ over 10 folds each); and the second one (labeled ``\emph{short-term}'') predicts the next image in the original photo album ($N{=}1742$). The problem is posed as a classification task choosing from the true image, and four other images selected uniformly at random from the same album. Here we also consider \textbf{NN} that simply picks the nearest neighbor, and \textbf{FI} that picks the furthest image from the given image, both in cosine distance of $\mathit{fc7}$ features. K-Means is not suitable for this task since it does not include temporal information. All methods were trained in an unsupervised manner for each concept as before.
\begin{figure}[t]
	\centering
	\includegraphics[width=.7\linewidth]{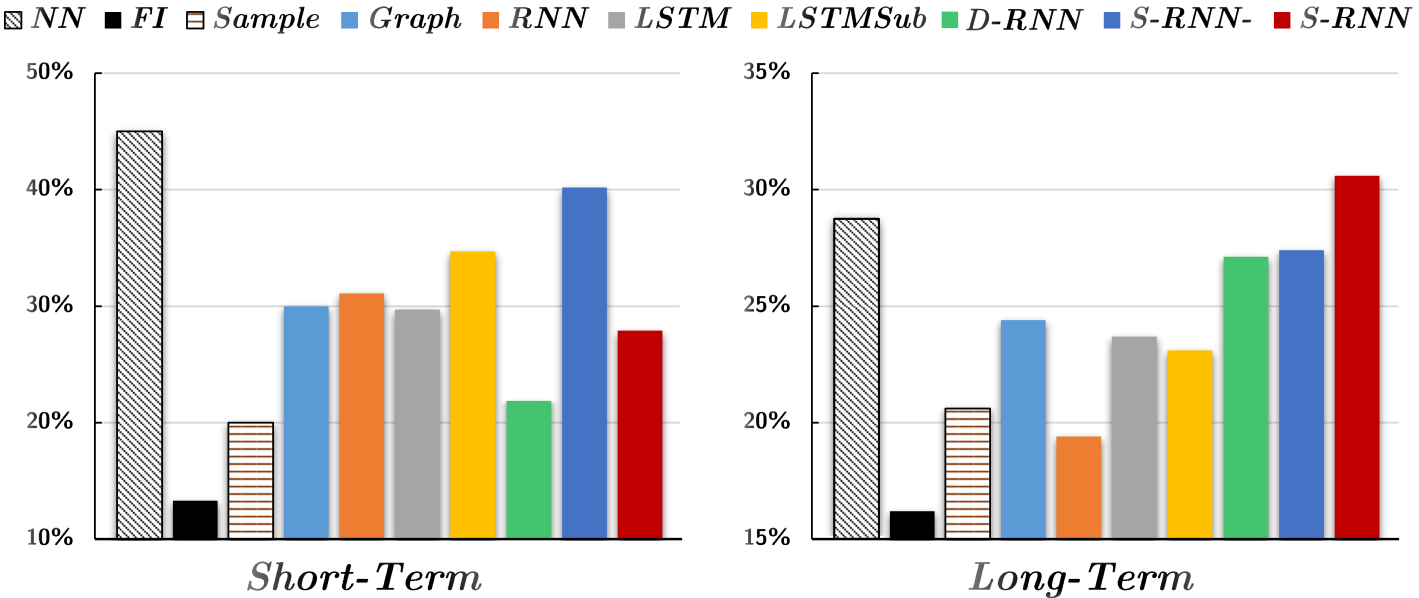}
	\caption{\emph{Predicting the next image}. S-RNN is best at capturing long-term correlations, and nearest-neighbors is best at capturing short-term correlations, as expected \label{fig:prediction}}
\end{figure}

\begin{figure}[tb]
    \centering
    \includegraphics[width=0.7\linewidth]{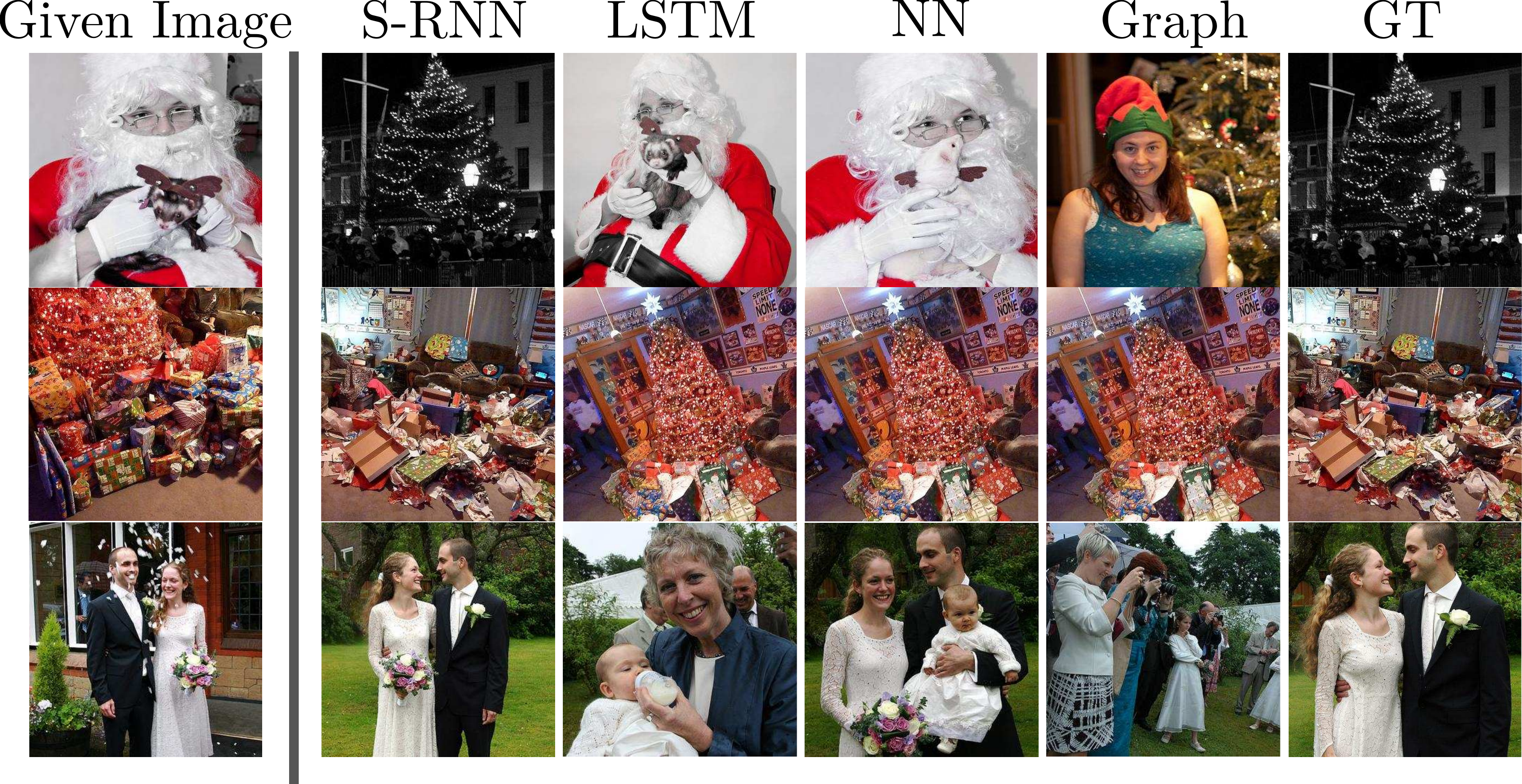}
    \caption{\textit{Long-term prediction}. Examples of the images predicted by our method compared to baselines. The image is chosen from a line-up of five images from the same album (generated by experts as summaries). We see our method captures \textit{Santa}$\rightarrow$\textit{Tree} and \textit{Closed Presents}$\rightarrow$\textit{Open Presents} while the baselines focus on similar images\label{fig:newcomparison}}
\end{figure}

\noindent \textbf{Results.} In Fig.~\ref{fig:prediction} we present results for the prediction of the next image. When we consider long-term interactions between images, {S-RNN} successfully predicts the next image in the storyline $\mathbf{31\%}$ of the time, significantly higher than baselines. On the other hand, we can see that when we simply want to predict consecutive images, NN is the best. To further visualize the results for ``\textit{long-term}'' correlations, we also give example comparisons with baseline methods in Fig.~\ref{fig:newcomparison}. 

\subsection{Task2: Photo Album Summarization\label{sec:sum}}

In the final experiment, we evaluate on the task of album summarization. In particular, we focus on summarizing an individual album based on the concept (\eg a \emph{Paris} album), rather than heuristics such as image quality or presence of faces~\cite{obrador2010supporting,sinha2011summarization,Sadeghi2015}. This experiment addresses the question whether storylines can help to summarize an album.

\noindent \textbf{Human Generated Summaries.} Photo album summarization is inherently a subjective and difficult task. To get a sense of the difficulty, we first compared the human summaries (used in Sec.~\ref{sec:imagepred}) to baselines with a separate AMT preference study. We had two findings. First, for some concepts, such as \emph{Wedding}, the albums are frequently already summaries by professional photographers, and thus generating summaries is trivial. Specifically, there is no significant difference between human generated summaries and uniformly sampling from the data distribution (Sample). We thus only evaluate on concepts where there is significant difference between human generated summaries and ones generated by baselines. Second, we found human generated summaries are only preferred $\mathbf{59.5\%}$ of the time against the strongest baseline. 

\noindent \textbf{Setup.} The photo albums for a given concept are randomly divided into a training set and a validation set with a ratio of 9:1, and no ground truth summaries were provided. We additionally consider the baseline \textbf{Local} where K-Means clustering is used for summarization by applying clustering on $\mathit{fc7}$ features for each individual album. As before, we assign the closest image to the cluster center for clustering-based methods. While it is not required for S-RNN, we sort the selected photos in temporal order as a post-processing step for all the baselines when necessary for fair comparison. 

\begin{figure}[tb]
	\centering
	\includegraphics[width=1.0\linewidth]{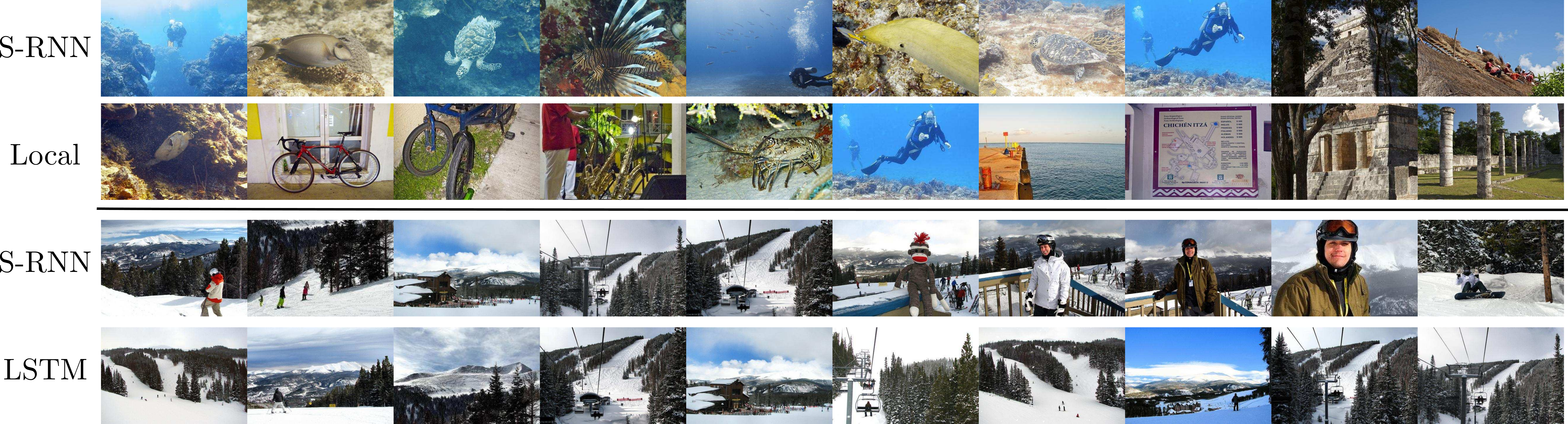}
	\caption{Examples of summaries generated by our method and two representative baselines for \emph{Scuba-diving} and \emph{Snowboarding}. In the \textit{Scuba-diving} example Local aims to capture diversity, and thus our method is more relevant. In \textit{Snowboarding}, LSTM focuses on short-term correlations, and chooses many similar images, while our method effectively captures the album\label{fig:summaryexamples}}
\end{figure}

\noindent \textbf{Qualitative Results.} The results for a few concepts are presented in Fig.~\ref{fig:summaryexamples}. We can see that S-RNN captures a set of relevant images without losing diversity. In contrast, Local captures only diversity, and LSTM that tries to learn short-term correlations between consecutive images, and as result often prefers similar images in a row. Additional summaries by {S-RNN} are presented in Fig.~\ref{fig:summaryteaser}. 
\begin{figure}[t]
	\centering
	\includegraphics[width=1.0\linewidth]{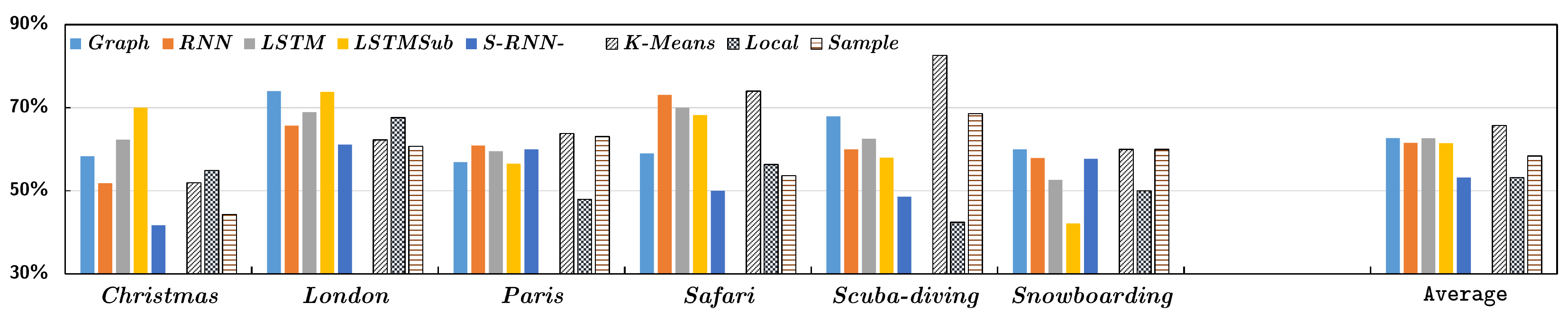}
	\caption{Photo album summarization. AMT pairwise preference between our method and multiple baselines. $70\%$ means that summaries by our method were preferred $70\%$ of the time. It is important to keep in mind that compared to the strongest baseline, a human generated summary was on average only preferred $59.5\%$ of the time. Sec.~\ref{sec:sum} contains a detailed explanation of the experiment setup and analysis of the results \label{fig:summarize}}
\end{figure}

\noindent \textbf{Quantitative Evaluation.} To directly compare the quality of the generated summaries, another AMT preference study was conducted. For {S-RNN} and each baseline, 200 random pairwise comparisons were generated. 
Each question was given to 5 separate workers for consistency. We used a consensus approach where a comparison gets a score of $1$ if there is a tie, or a score of $2$ if there is consensus.

In Fig.~\ref{fig:summarize} we present comparison with the baselines. We can see that on average our method is preferred over all the baselines. To provide a more detailed analysis, we divide the baseline methods into two groups: the \emph{Storyline} group (filled with pure colors) that captures the latent temporal information in the data, and the {Non-Storyline} group (filled with patterns) that do not. The \emph{Storyline} group includes Graph, RNN, LSTM, LSTMSub, and S-RNN- (Our method also falls into this group), while the \emph{Non-Storyline} group has K-Means, Local and Sample. There are few interesting points:
\begin{enumerate}
    \item S-RNN performs relatively better on travel-related albums (\emph{Paris}, \emph{London}) suggesting it is easier to latch onto landmarks than high-level concepts like in \emph{Christmas}.
    \item For concepts like \emph{Christmas}, methods that learn short-term correlations from the data distribution are still preferred by the users. The fact that S-RNN- outperforms LSTMs and RNNs, can be interpreted as follows. RNNs suffer from the curse of dimensionality if naively applied to storyline learning, but the S-RNN loss reduces the dimensionality of the output space by an order of magnitude (4096 to 100s).
    \item While simple as they seem, Local and Sample are very competitive baselines. We believe the reason is that Local aims to provide a diverse set of images from each album, and Sample is representative of the underlying data. Therefore, with the post-processing step that re-arranges the selected images in temporal order, these methods can do well on good albums. However, they do poorly when the album is noisy, as illustrated in Fig.~\ref{fig:summaryexamples} first example.
\end{enumerate}

\noindent \textbf{Does Time Information Help Summarization?} For further analysis, we compared the described S-RNN with S-RNN trained on shuffled data (ordering discarded) with a preference study on AMT. S-RNN using the time information was preferred $68.4\%$ over S-RNN without time information, demonstrating that the time information significantly helps to generate a summary liked by people.

\begin{figure}[t]
	\centering
	\includegraphics[width=1.0\linewidth]{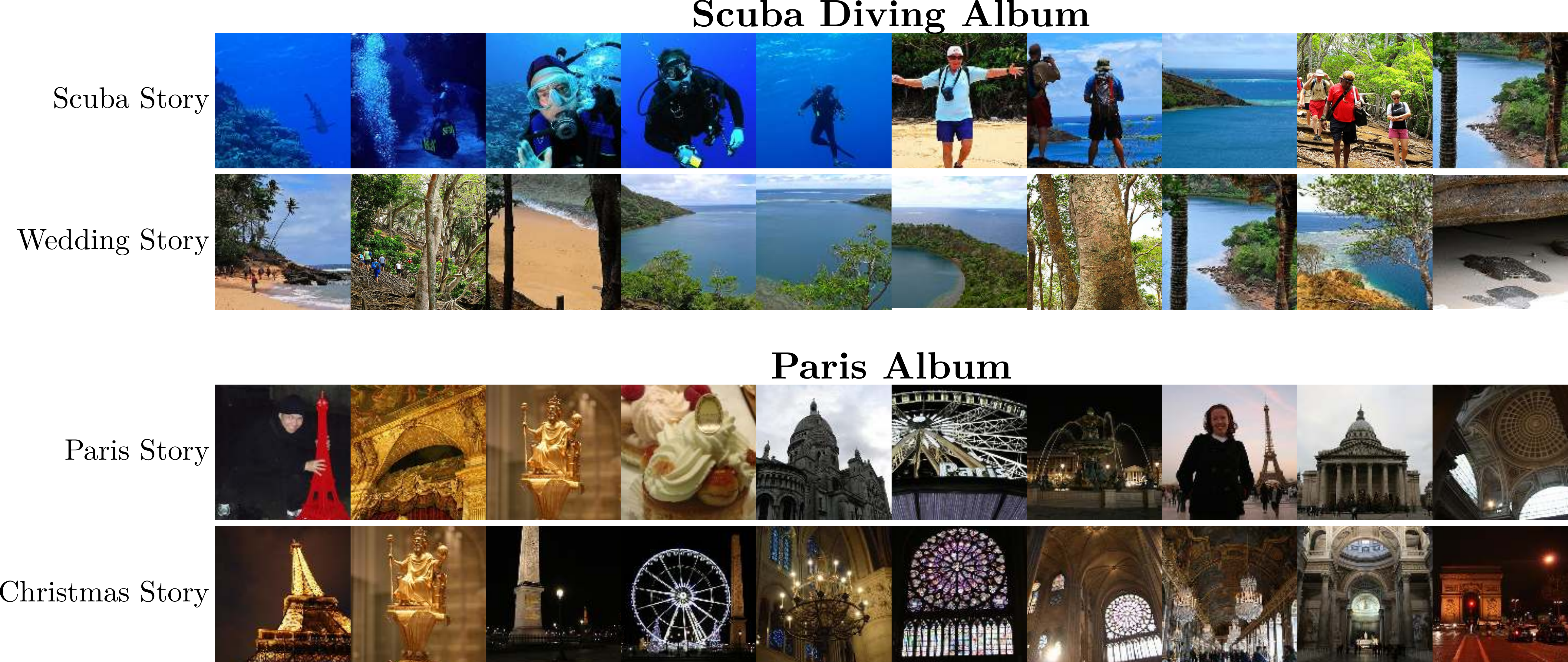}
	\caption{The first two rows show a \emph{Scuba Diving} album summarized with a \emph{Scuba} model and a \emph{Wedding} model, and the last two show a \emph{Paris} album summarized with a \emph{Paris} model and a \emph{Christmas} model. The \emph{Wedding} story emphasizes the beach resort images of the \emph{Scuba} album, and the \emph{Christmas} story emphasizes the churches and sparkling lights images in the \emph{Paris} album \label{fig:diff}}
\end{figure}

\noindent \textbf{Transferring storyline knowledge.} Each album can have different stories and themes. In Figure~\ref{fig:diff} we present two different summaries of two photo albums. The first album is a \emph{Scuba Diving} album, and the first summary from that album is generated with the model trained on \emph{Scuba Diving} albums. In the second row, the same album is summarized using a model trained on \emph{Wedding} albums. We can see that this emphasizes scenic beach pictures reminiscent of a beach resort wedding. The second album is a \emph{Paris} album, and the first summary is generated using \emph{Paris} model. The second summary however, is generated using a \emph{Christmas} model, and we can see that this emphasizes pictures of churches and sparkling lights at night.

\section{Conclusion}

We have presented an approach to learn visual storylines for concepts automatically from the web. Specifically, we use Flickr albums and train an S-RNN model to capture the long-term temporal dynamics for a concept of interest. The model is designed to overcome the challenges posed by high correlations between consecutive photos in the album if sequence predictors are directly applied. We evaluate our model on learning storylines, image prediction and album summarization, and show both qualitatively and quantitatively that our method excels at both extracting salient visual signals for the concept, and learning long-term storylines to capture the temporal dynamics.
 
\section{Acknowledgements}
This research was supported by the Yahoo-CMU InMind program, 
ONR MURI N000014-16-1-2007,
and a hardware grant from Nvidia. The authors would like to thank Olga Russakovsky and Christoph Dann for invaluable suggestions and advice, and all the anonymous reviewers for helpful advice on improving the manuscript.


\bibliographystyle{splncs}
\bibliography{storylines}

\newpage
\section{Appendix}

\subsection{EM-Derivation of the Update Equations}

We begin with a standard RNN.
Given $T$ data points $\xbf_{1:T}=\left\{x_1,\ldots,x_{T}\right\}$ (images) and the model parameters $\mcm$, RNN maximizes $\bPf\left(x_{t}|\xbf_{1:t-1};\mcm\right)$ at step $t$, which is an exact decomposition of $\bPf(\xbf_{1:T};\mcm)$, no independence assumed.

The key idea of S-RNN is to train over all ordered subsets (of size $N$) in $\xbf_{1:T}$, selected by latent variables ${\fzf_{1:N}{=}\left\{z_1,\ldots,z_{N}\right\}}$. Here we assume the likelihood of $\xbf_{1:T}$ selected by $\fzf_{1:N}$, is only related to the selected subset ${\hat{\fzf}_{1:N}{=}\left\{\hat{z}_1,\ldots,\hat{z}_{N}\right\}}$, namely (here $\xbf_{\hat{\fzf}_{1:N}}{=}\left\{x_{\hat{z}_1},\dots,x_{\hat{z}_N}\right\}$):
\begin{equation}
    \label{eq:assume2}
    \hspace{-4mm}\bPf(\xbf_{1:T}|\fzf_{1:N}{=}\hat{\fzf}_{1:N};\mcm) \propto \bPf(\xbf_{\hat{\fzf}_{1:N}}|\fzf_{1:N}{=}\hat{\fzf}_{1:N};\mcm).
\end{equation}

\noindent S-RNN maximizes the likelihood of $\xbf_{1:T}$ over all $\fzf_{1:N}$:
\begin{align}
    \max_\mcm \bPf(\xbf_{1:T};\mcm) &= \max_\mcm \sum_{\fzf_{1:N}} \bPf(\xbf_{1:T}|\fzf_{1:N};\mcm)\bPf(\fzf_{1:N}) \\
    &=\max_\mcm \sum_{\fzf_{1:N}} \bPf(\xbf_{\fzf_{1:N}}|\fzf_{1:N};\mcm) \bPf(\fzf_{1:N}) \label{eq:srnn} \\
    &=\max_\mcm \sum_{\fzf_{1:N}} \left( \prod_n \bPf(x_{z_{n+1}}|\xbf_{\fzf_{1:n}},\fzf_{1:N};\mcm) \right) \bPf(\fzf_{1:N}) \label{eq:wrong}
\end{align}
Here we assume the prior $\bPf(\fzf_{1:N})$ does not depend on $\mcm$, and use Eq.~\ref{eq:assume2}.
In Eq.~\ref{eq:wrong} we also used the chain rule to make it more similar to RNN (this is Eq. 4 in the paper), but here 
Eq.~\ref{eq:srnn} is directly solved with the EM-algorithm and factorized later. In the {\bf E-Step}, we sample $\fzf$ to approximate the expectation: (for simplicity we remove the subscripts in $\fzf_{1:N}$ and $\hat{\fzf}_{1:N}$)
\begin{align}
    Q(\mcm;\mcm_0) := \bEf_{\hat{\fzf}\sim q_0}\left[\log \left( \bPf(\xbf_{\hat{\fzf}}|\fzf{=}\hat{\fzf};\mcm)\bPf(\fzf{=}\hat{\fzf}) \right) \right], \label{eq:sample2}
\end{align}
where $q_0$ is $\bPf(\fzf|\xbf;\mcm_0)$. 
For a single sample $\hat{\fzf}_{1:N}$, the {\bf M-Step} follows: \\
(for simplicity of notation $\bPf(\xbf_{\hat{\fzf}}|\fzf{=}\hat{\fzf};\mcm) {=} \bPf(\xbf_{\hat{\fzf}};\mcm)$)
\begin{align}
    \max_\mcm Q(\mcm;\mcm_0) 
    &=\max_\mcm \log \bPf(\xbf_{\hat{\fzf}_{1:N}};\mcm). \\
    &=\max_\mcm \sum_n \log \bPf(x_{\hat{z}_{n+1}}|\xbf_{\hat{\fzf}_{1:n}};\mcm). \label{eq:m2}
\end{align}
This is the standard RNN objective except over a subset.
The final implementation detail, is that we can rewrite $\bPf(\fzf|\xbf;\mcm_0)$ in a simpler form:
\begin{align}
    \bPf(\fzf{=}\hat{\fzf}|\xbf;\mcm_0) &\propto \bPf(\xbf|\fzf{=}\hat{\fzf};\mcm_0)\bPf(\fzf{=}\hat{\fzf};\mcm_0) \\
    &\propto \bPf(\xbf_{\hat{\fzf}}|\fzf{=}\hat{\fzf};\mcm_0)\bPf(\fzf{=}\hat{\fzf}) \\
    &\propto \bPf(\xbf_{\hat{\fzf}};\mcm_0)\bPf(\fzf{=}\hat{\fzf}) \\
    &= \prod_n \bPf(x_{\hat{z}_{n+1}}|\xbf_{\hat{\fzf}_{1:n}};\mcm_0) \bPf(z_{n+1}{=}\hat{z}_{n+1}|\fzf_{1:n}{=}\hat{\fzf}_{1:n}) \label{eq:sample3}
\end{align}
where we used Bayes rule, Eq.~\ref{eq:assume2}, and the chain rule of probability. This equation allows us to sample $z$ sequentially, and train the RNN as before. $\bPf(z_{n+1}|\fzf_{1:n})$ is just the sequential version of the prior $\bPf(\fzf_{1:N})$ and captures the fact that $\fzf$ defines an ordered subset. ($z_n \in \{ 1,2,\dots,T \}$, and $z_n < z_{n+1}$)

In summary, the training method simply alternates sampling from $\bPf(\fzf|\xbf;\mcm_0)$ (E-Step in Eq.~\ref{eq:sample2}) and updating $\mcm$ using Eq.~\ref{eq:m2} (M-Step). This falls neatly into the RNN pipeline, with only a simple sampling step before feeding new samples to the network. 
Pseudo-code for training S-RNN is presented in Algorithm~\ref{alg:code}, for a single training sequence. Note that $\bPf(x {=}x_j | \xbf_{\hat{\fzf}_{1:n}};\mcm)$ is the prediction of the RNN at step $n$, and $\mathrm{Cat}$ is the categorical distribution. The RNN is updated by using the prediction of the model at step $n$ ($\mathbf{y}_n$ in the paper), $x_{\hat{z}_{n+1}}$, and $\mathcal{X}_n$, with the loss function (described in Eq.~5 the paper). This provides a gradient which is back-propagated through the network.

Note that the sampling depends on the model prediction. Intuitively, this is the key that allows exploring all subsets of the data, since the model is used to intelligently guide the exploration.

\begin{algorithm}
  \caption{
    \label{alg:code}}
  \begin{algorithmic}[1]
    \Function{Train S-RNN}{$\{x_1,x_2,\dots\}$} \Comment{Images}
      \State $\hat{\fzf}_{1:N},\mcm \gets \mathrm{random}$ \Comment{Randomly initialize}
      \State $n \gets 1$ 
      \While{$\textrm{not converged}$}
        \State $\mathcal{X}_n \gets \{ x_{{z_n}+1}, x_{{z_n}+2}, \dots \}$ \Comment{Set of future images}
        \State $p_j \gets 0$
        \For{$x_j \in \mathcal{X}_n$}
          \State $p_j \gets \bPf(x {=}x_j | \xbf_{\hat{\fzf}_{1:n}};\mcm)\bPf(z_{n+1}{=}j|\fzf_{1:n}{=}\hat{\fzf}_{1:n})$
        \EndFor
        \State $\hat{z}_{n+1} \sim \mathrm{Cat}({p_1,p_2,\dots})$ \Comment{Sample $\hat{z}_{n+1}=j$ with probability $p_j$}
        \State $\textrm{Update RNN (} \mcm \textrm{) using } x_{\hat{z}_{n+1}},\mathcal{X}_n$ \Comment{See text}
        \State $n \gets n+1$
      \EndWhile
      \EndFunction
  \end{algorithmic}
\end{algorithm}

\subsection{Choosing the size of the summary}

\begin{figure}[t]
\centering
\includegraphics[width=0.6\linewidth]{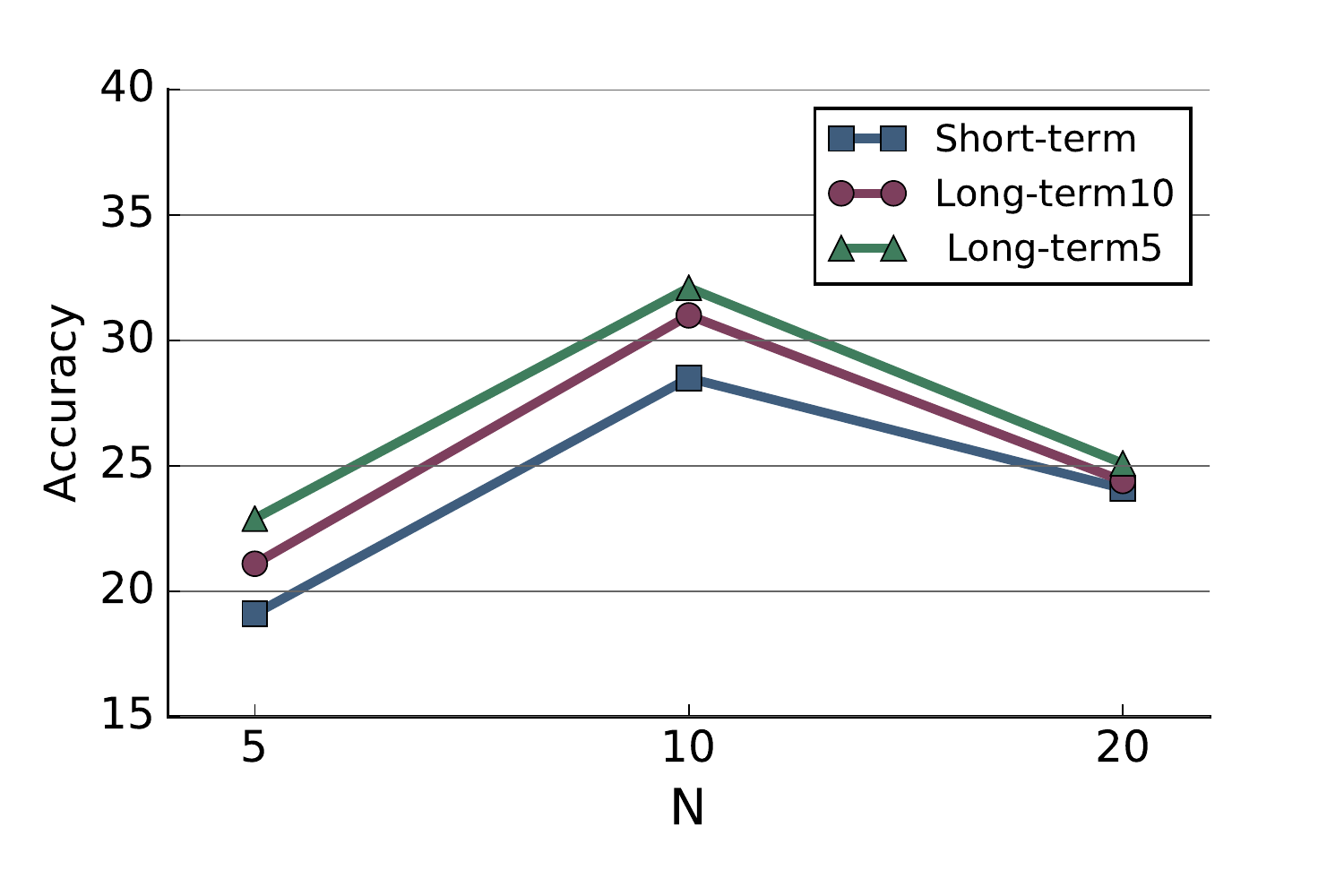}
\caption{The performance for S-RNN with three values of $N=5,10,20$ on three tasks: Short-term prediction (Short-term), Long-term prediction as before (Long-term10), and Long-term prediction using a ground truth of size 5 instead of 10 (Long-term5).}
\label{fig:size_of_n}

\end{figure}

In this work, we fixed the size of the storyline and summaries to be $N=10$ for all the concepts as a good compromise between content and brevity, i.e. intuitively allows for a compact but informative summary of a concept. 
In Fig.~\ref{fig:size_of_n} we follow the same setup as in Section~4.4 in the paper. We plot the performance for S-RNN with three values of $N=5,10,20$ on three tasks: Short-term prediction (Short-term), Long-term prediction as before (Long-term10), and Long-term prediction using a ground truth with storylines of length 5 (Long-term5). We observe that the model trained with $N=10$ has the highest performance on all three tasks. This implies that S-RNN is not overfitting to only the case of 10 image storylines, since then we would expect the $N=5$ method to do significantly better for Long-term5.
Our interpretation is that for small N, the sequences are short and easy to learn, but not very informative. For large N, the sequence approaches the full album, and more data and model capacity is needed to learn the long-term correlations.

\subsection{Dataset}

\begin{figure}[b]

\centering
\includegraphics[width=0.7\linewidth]{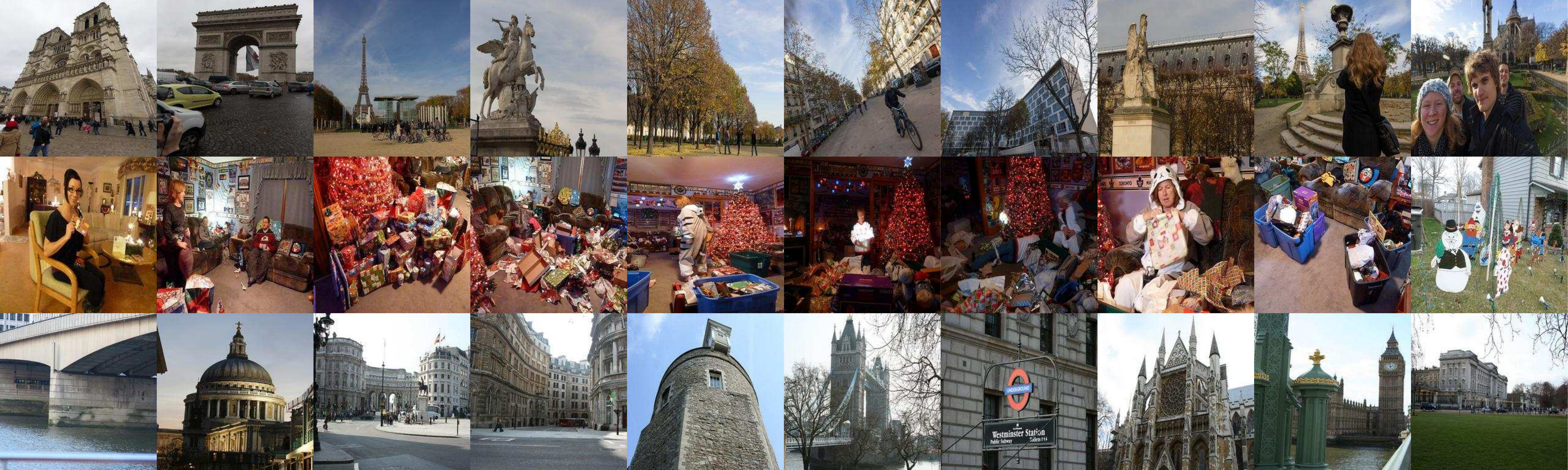}
\caption{Example human-generated summaries for \textit{Paris}, \textit{Christmas}, and \textit{London} albums.}

\label{fig:dataset2gt}
\end{figure}

In Fig.~\ref{fig:dataset2} we sample images from two randomly selected albums for each concept. As we can see, some albums are quite relevant to the concept, such as the first \textit{Wedding} album. However for this simple sampling, many albums, such as the second \textit{Christmas} and first \textit{Safari} albums, have little relevance. This is typical of the dataset, since a simple query for \textit{Christmas} returns, for example, Christmas parades, Christmas plays, and community events, in addition to the expected family celebrations. Moreover, it can be seen that many images are not related to the concept. What this means for our tasks, is that retrieving relevant images for a concept is difficult. However, in any given album, if the album is good, then uniformly sampling images does quite well at summarizing that album. In addition, in Fig.~\ref{fig:dataset2gt} we show few human-generated summaries for the albums.

\subsection{Storylines for Different Concepts}

\begin{figure}[htbp]
\centering
\includegraphics[width=0.9\linewidth]{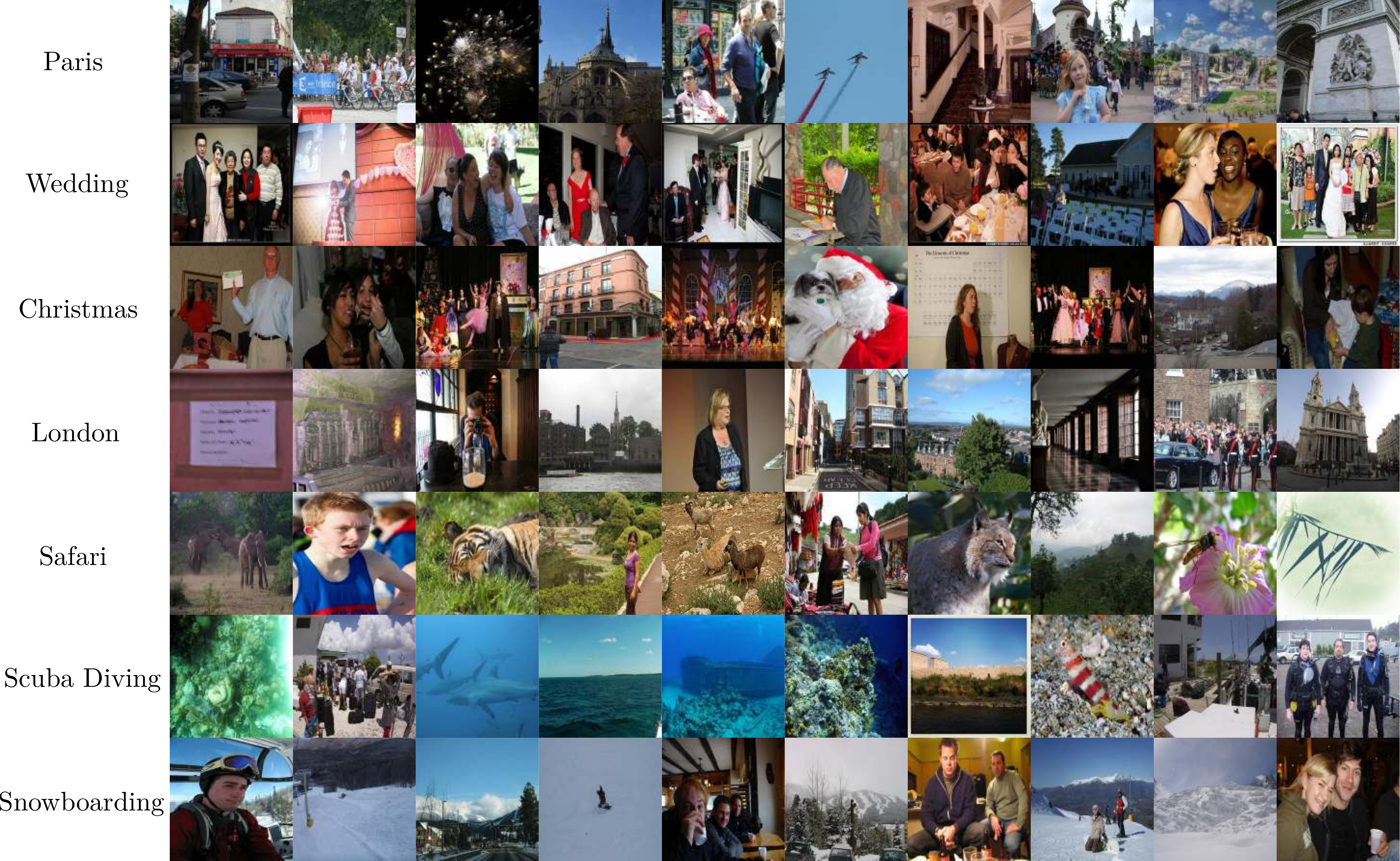}
\caption{(\textbf{K-Means}). Storylines by K-Means for different concepts. Each collection contains some relevant images, but many of the images are not relevant, since K-Means emphasizes diversity. Furthermore, each collection is not very coherent.}
\label{fig:kmeansmega}
\end{figure}

In Fig.~\ref{fig:kmeansmega}, we motivate the problem of selecting images for the storyline by presenting, for each concept, K-means clusters using $fc7$ features. We see that while some images are relevant in each group, some concepts are very different from what we expect, for example it is difficult to recognize \textit{Paris} or \textit{London}. We argue that perhaps more importantly, all the images in the collection do not complement each other, that is, there is no sequence of events, or relationship between the images.

\begin{figure}[htbp]
\centering
\includegraphics[width=0.9\linewidth]{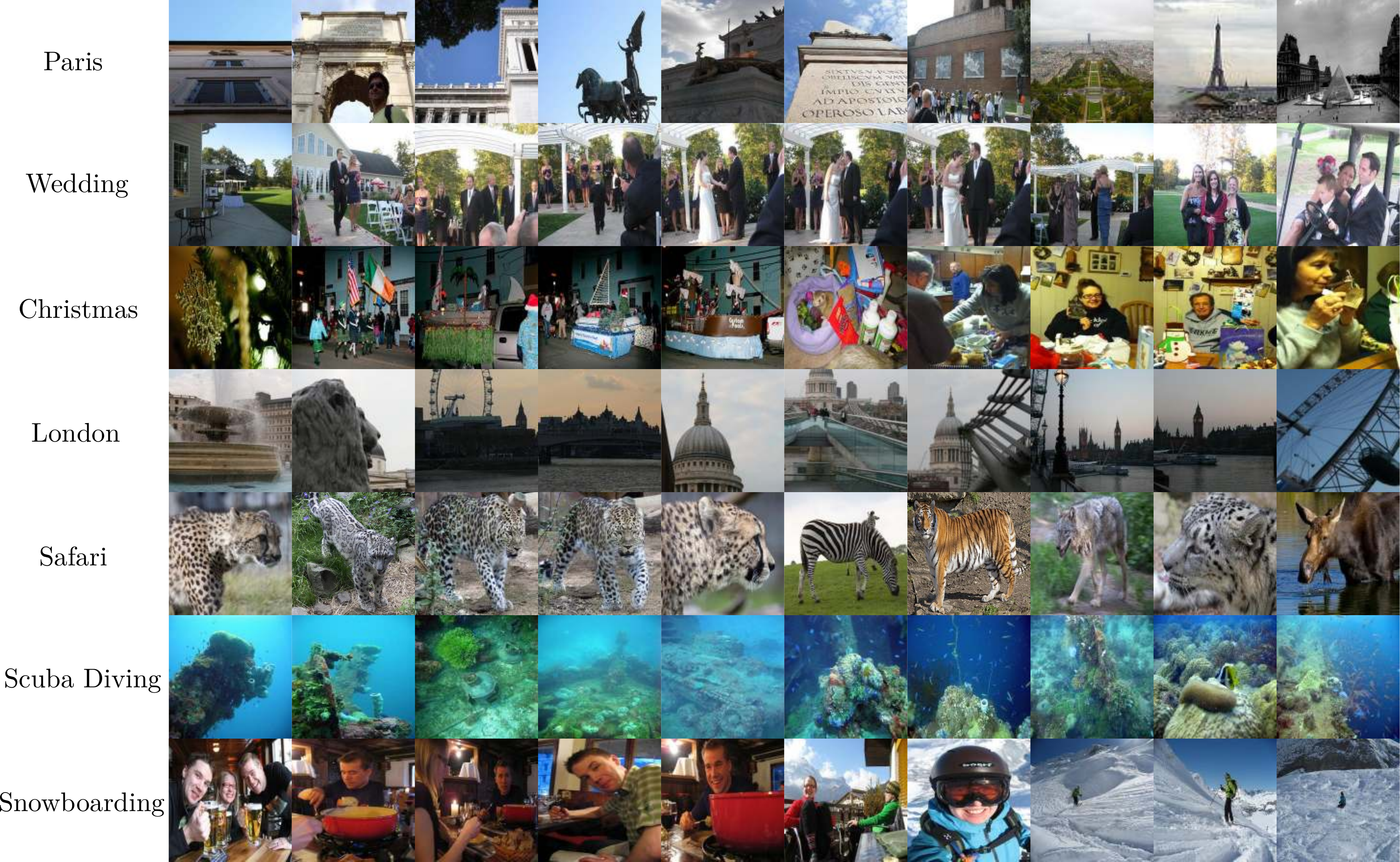}
\caption{(\textbf{S-RNN}) Storylines by our method for different concepts. Each collection of images is more related to the concept, and each collection captures a coherent story of the concept.}
\label{fig:srnnmega}
\end{figure}

\begin{figure}[htbp]
\centering
\includegraphics[width=0.9\linewidth]{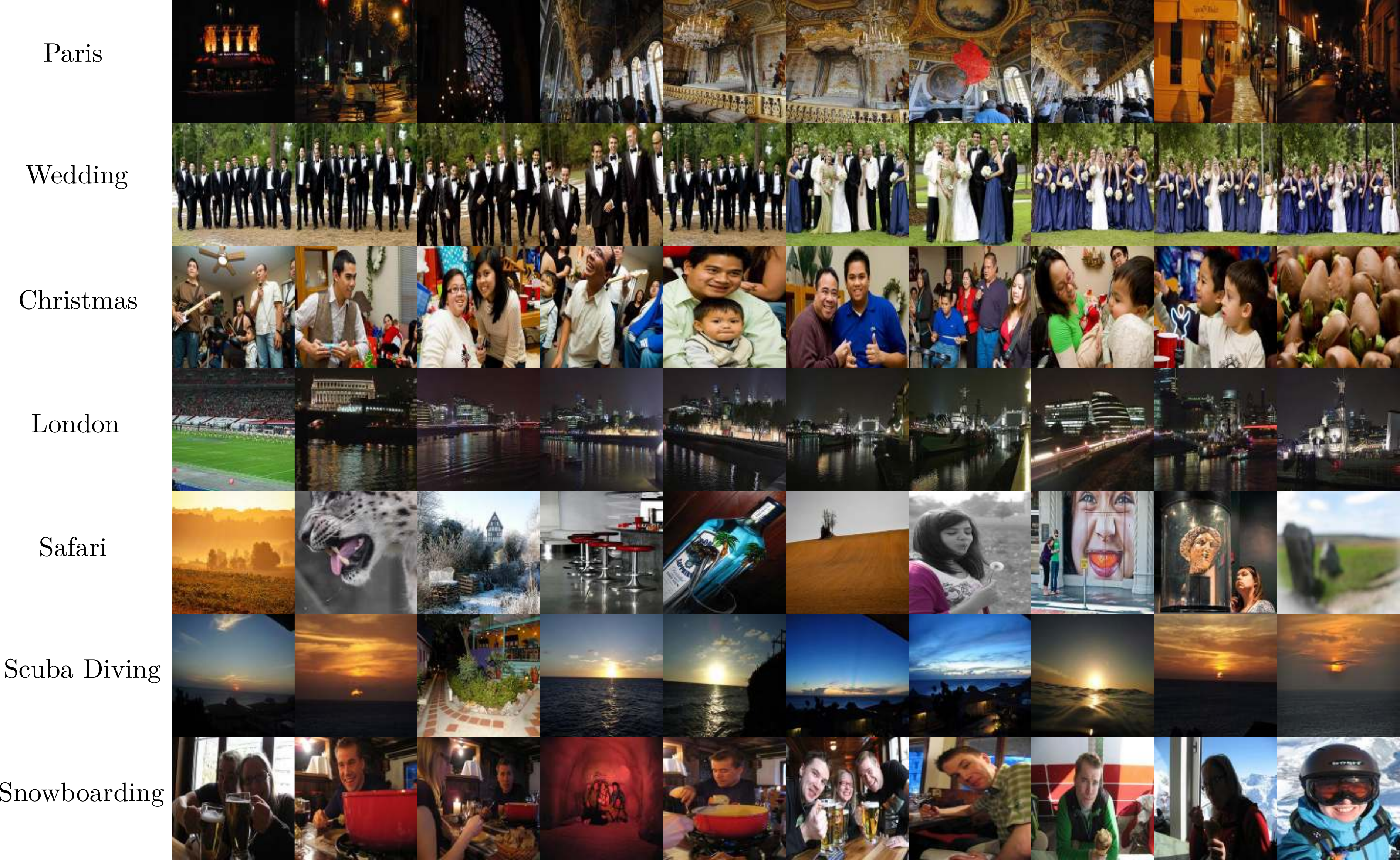}
\caption{(\textbf{S-RNN-}). Storylines by our method without skipping for different concepts. While these collections are relevant to the topic, and the collections are coherent, it appears that the model is focusing on short-term correlations, and the images are not very diverse. }
\label{fig:srnnmmega}
\end{figure}

However, in Fig.~\ref{fig:srnnmega} we present the images retrieved by our method (\textbf{S-RNN}). We can see that each collection (each row), captures a coherent story of the concept. If we compare that with our method without skipping (\textbf{S-RNN-}, Fig.~\ref{fig:srnnmmega}), we see that the coherence in those collections is focused on short-term correlations, and those baselines choose highly correlated images.

\subsection{Example Summaries}

\begin{figure}[t]
\centering

\includegraphics[width=0.85\linewidth]{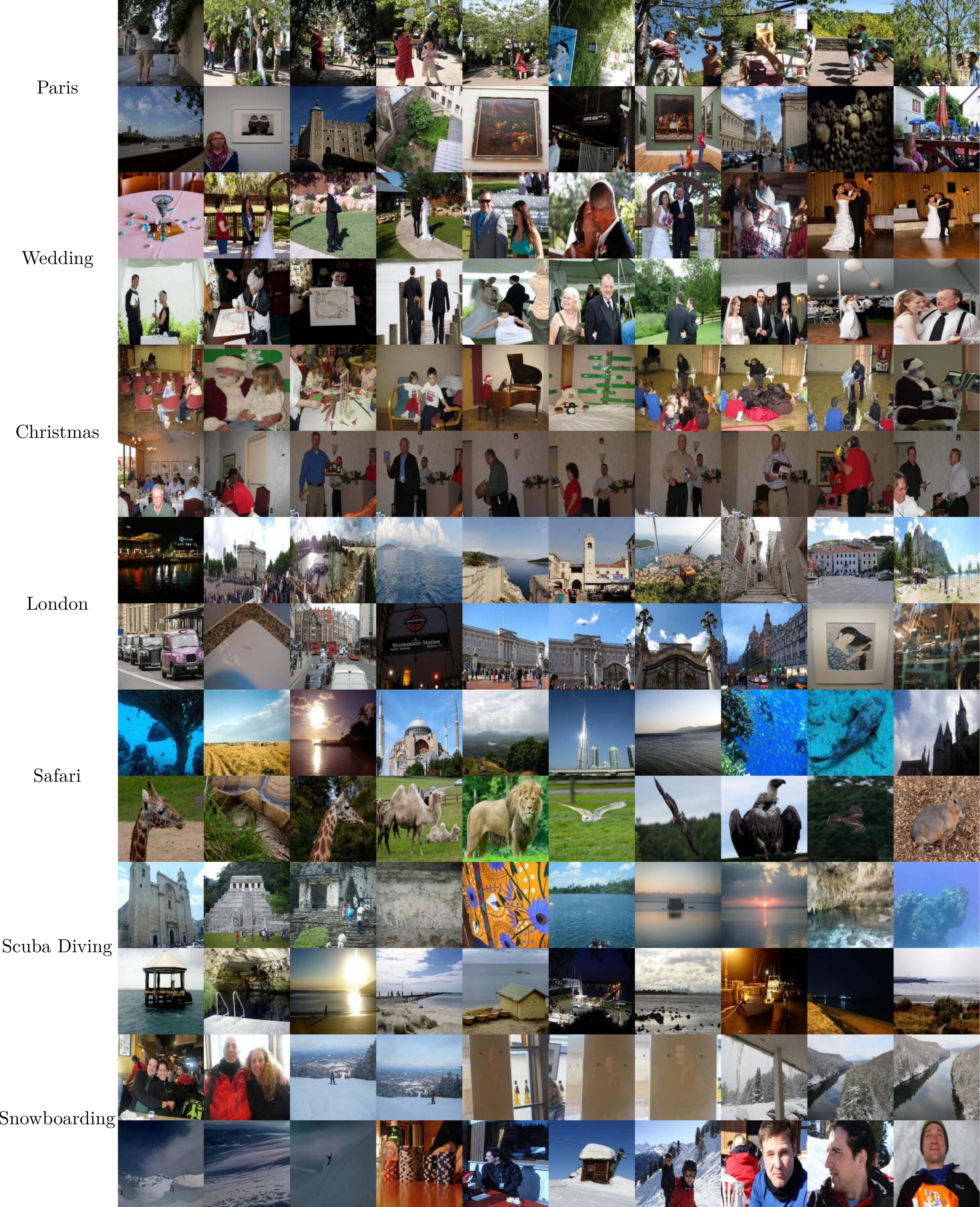}
\caption{Each row consists of a randomly selected album from the dataset, two for each topic. Each album is visualized using 10 images uniformly sampled from the album.}

\label{fig:dataset2}
\end{figure}

\begin{figure}[htbp]
    \centering
    \includegraphics[width=1.0\linewidth]{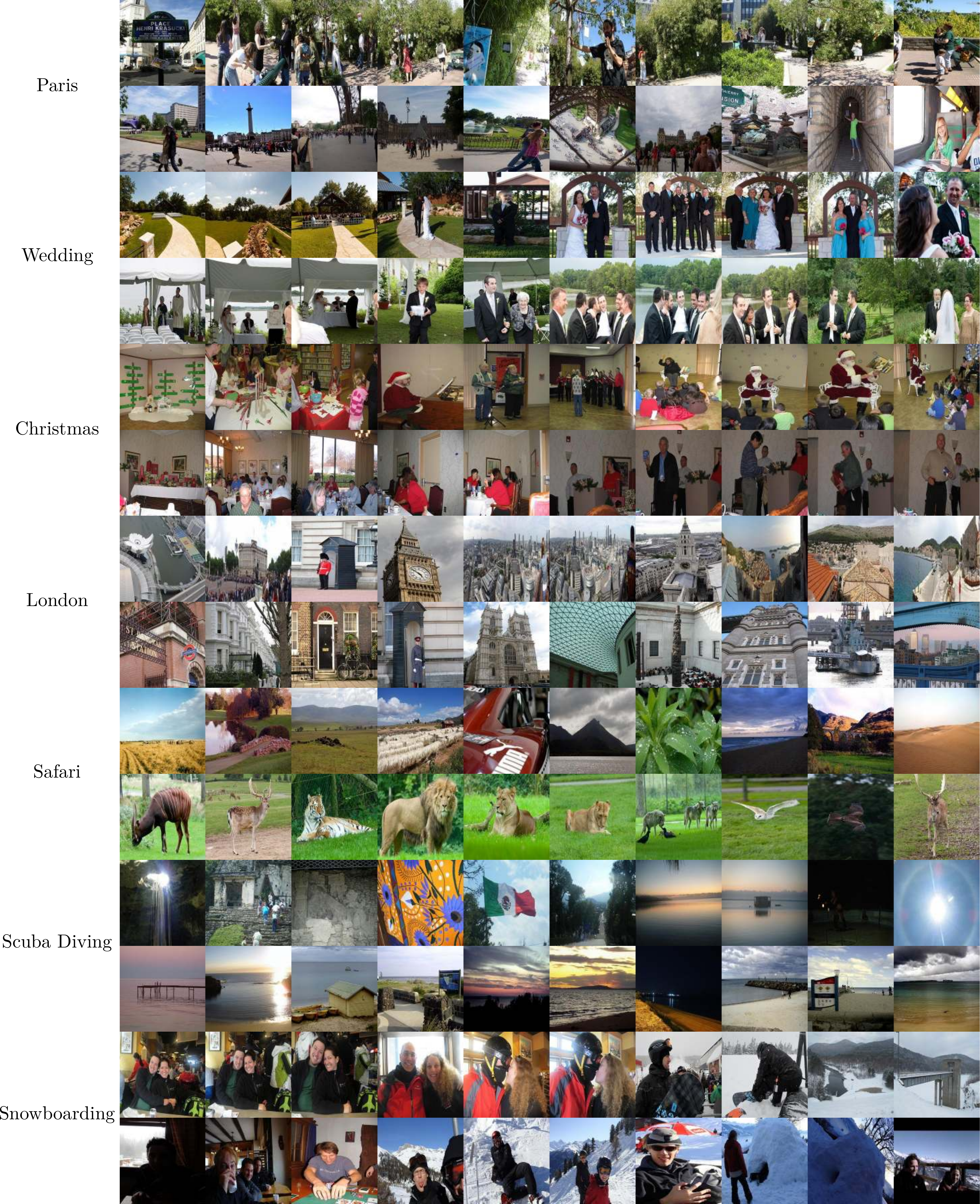}
    \caption{Randomly selected summaries using our method. These albums correspond to the albums presented in Fig.~\ref{fig:dataset2}. It is difficult to judge the quality of the summaries without seeing the hundreds of images behind each summary, but hopefully provide insight.}
    \label{fig:examples}
\end{figure}

We present many randomly selected summaries from the dataset. Photo album summarization depends heavily on the quality of the given album, and the album's relevance. Summarization is therefore a more difficult task than selecting relevant images for evaluation, since on some albums, simple baselines do remarkably well. In Fig.~\ref{fig:examples} we present summaries by our method (\textbf{S-RNN}) on the same randomly selected albums as in Fig.~\ref{fig:dataset2}.

\newpage


\end{document}